\documentclass[11pt]{article}
\usepackage{coling2018}
\usepackage{times}
\usepackage{url}
\usepackage{latexsym}
\usepackage{url}
\usepackage{amssymb}
\usepackage{amsmath}
\usepackage{latexsym}
\usepackage{graphicx}
\usepackage{algpseudocode}
\usepackage{algorithm2e}
\usepackage{enumitem}
\usepackage{breqn}
\usepackage{multirow}
\usepackage{slashbox}
\usepackage{adjustbox}

\title{Leveraging Medical Sentiment to Understand Patients Health on Social Media}

\author{Shweta Yadav$^{\ast}$, Joy Sain$^{\dagger}$, Amit Sheth$^{\dagger}$, Asif Ekbal$^{\ast}$, Sriparna Saha$^{\ast}$, Pushpak Bhattacharyya$^{\ast}$ \\
  $^{\ast}$Indian Institute of Technology Patna, India  \\
 $^{\dagger}$Kno.e.sis Center, Wright State University, USA \\
  {\tt $^{\ast}$\{shweta.pcs14,asif,sriparna,pb\}@{iitp.ac.in}, $^{\dagger}$amit@knoesis.org}}

\date{}

\begin{document}
\maketitle
\begin{abstract}
 The unprecedented growth of Internet users in recent years has resulted in an abundance of unstructured information in the form of social media text. A large percentage of this population is actively engaged in health social networks to share health-related information. In this paper, we address an important and timely topic by analyzing the users' sentiments and emotions w.r.t their medical conditions. Towards this, we examine users on popular medical forums (\url{Patient.info, dailystrength.org}), where they post on important topics such as \textit{asthma}, \textit{allergy}, \textit{depression}, and \textit{anxiety}. First, we provide a benchmark setup for the task by crawling the data, and further define the sentiment specific fine-grained medical conditions (\textit{Recovered, Exist, Deteriorate}, and \textit{Other}). 
We propose an effective architecture that uses a Convolutional Neural Network (CNN) as a data-driven feature extractor and a Support Vector Machine (SVM) as a classifier. We further develop a sentiment feature which is sensitive to the medical context. Here, we show that the use of medical sentiment feature along with extracted features from CNN improves the model performance. In addition to our dataset, we also evaluate our approach on the benchmark \textit{``CLEF eHealth 2014"} corpora and show that our model outperforms the state-of-the-art techniques. 

\end{abstract}
\section{Introduction}
\label{intro}

The phenomenal rise in blogging is coupled with the increasing popularity of medical forums for sharing medical problems or experiences, and seeking for health-related information or opinions of other users (i.e., patients or health professionals).
According to a recent study conducted by the Pew Internet \& American Life Project\footnote{http://www.pewinternet.org/}, almost 80\% of Internet users in the US have explored health-related topics online. More often, 63\% of people look for the information about specific medical problems and nearly 47\% of the users search for the medical treatments or procedures over the Internet. \\
These self-narrated journals consist of a diverse variety of information, including,
user-specific concerns, triggers, reactions,
or merely status updates on their emotional states.
With this ever-increasing size of the blogs, a majority of these blog posts are left unused or unanswered. 
Considering this fact, it would be helpful to have a sentiment analyzer that could study the user's sentiment associated with the post related to his/her health status. Moreover, extracting sentiment and/or opinions from medical text can be crucial to assess the patient's health and also to help health professionals with automated decision support system. 
In this paper, we make an attempt to capture \textit{medical sentiment} (MS) from unstructured text by analyzing the subjectivity expressions describing a patient's medical conditions to prioritize the individual’s posts that requires immediate attention. \\
The existing research in sentiment analysis is primarily focused on detecting users' external sentiment \cite{shickel2016self} towards some entities like products, organizations, or events etc. In contrast, when working with the self-narrative medical journals (blogs), the focus is more on gauzing the users' internal sentiments towards their own emotions, feelings, and thoughts. The difference in objectives, sentiment types, and distribution of polarity introduce unique challenges for applying traditional sentiment analysis to this new domain. \\
One key aspect that differentiates our problem from the traditional sentiment analysis is our way of defining the polarity class. Traditionally, the sentiment is classified as either positive, negative, or neutral. 
In contrast, the notion of sentiment in the medical context is more granular which can be studied after considering various aspects \cite{denecke2015sentiment} that can directly impact the users' health conditions as studied by \cite{yadav2018multi} such as:
\begin{itemize}
\item \textbf{Changes in the medical condition} (example, Sentiment can be observed as a change in a patient's medical condition which can improve or worsen over a time.)
\item \textbf{Severity of the medical condition that impacts patient life} (example, \textit{severe headache impacts the patient's life more than a mild headache.})
\item \textbf{Outcome of a treatment} (example, there may be positive or negative impacts on a patient's treatment.)
\end{itemize} 
According to our analysis, nearly 95\% of the user-posts on medical forums carry negative sentiment towards their medical conditions. The negative sentiment alone is not informative enough for health professionals to make any clinical decision. Considering this fact, we further divide the negative sentiment specific class into \textit{`Exist'} and \textit{`Deteriorate'}, which explicitly conveys whether a user is experiencing a disorder or the existing has worsened over the time. We also analyze the positive sentiment of a user and project it as 'Recovered', conveying recovery from the disorders. \\
Although, several techniques exist to capture the sentiments in general domains, the sentiments expressed in medical narratives have not been analyzed and exploited well in required measures yet. The research in medical sentiment analysis is primarily focused on biomedical literature and Electronic Medical Record (EMR) documents. Recently, preliminary studies \cite{shickel2016self,yang2016mining} have been conducted to understand the sentiment in the medical setting. Several shared tasks \cite{losadaclef,CLPsych:2017} have also been organized to study the patient health-related opinions on social media. However, a majority of these studies are focused on mental health disorder, and the defined classification schemes are framed to understand the depressive behavior of a user using PHQ-9 levels \cite{kroenke2001phq} (a method to monitor the severity of depression).


\textbf{Problem Statement:} For a given medical blog post $M$, consisting of $n$ sentences i.e. $M = \{s_1, s_2, s_3,.....s_n\}$, the task is to predict the medical sentiment specific label `$y$' from a discrete set of medical condition `$Y$=\{\textit{Recovered, Exist, Deteriorate, Other}\}'. \\
We started with the Convolutional Neural Network (CNN) \cite{lecun1995convolutional} given the recent success in several Natural Language Processing (NLP) tasks \cite{kim2014convolutional,collobert2011natural,kalchbrenner2014convolutional,mikolov2013efficient,collobert2011natural,ekbal2016deep}.
Many applications use CNN for having automatic feature extraction capability. However, as we explain in Section $3$, the analysis of data has provided an interesting insight: \textit{while CNN-extracted features (making use of word embedding) capture the semantic information, the incorporation of external features could further assist in precisely capturing the subjectivity (sentiment, emotion) associated with medical concepts.} Correspondingly, we devise Medical Sentiment-CNN (MS-CNN) to advance conventional CNN by embedding various medical sentiment features. Specifically, we use a Support Vector Machine (SVM) \cite{cortes1995support} as a strong classifier instead of the softmax (Logistic Regression) classifier at the output layer of the CNN because of the limitation of logistic regression (LR). This is informed by the observation that when the data is non-linearly separable, an SVM with a non-linear kernel outperforms LR \cite{pochet2006support}.\\
\textbf{Contributions}: (i) a description of the medical sentiment analysis task by mining medical blog using crowd intelligence, (ii) development of an annotated medical sentiment corpus to the research community, and (iii) a method for online rapid assessment of medical sentiment with the fusion of CNN-generated features and the sentiment-sensitive medical features learned on an SVM to significantly improve the accuracy.

Our evaluations show that compared to baseline system (SVM \& CNN), our proposed approach  yields 15.27\% and 4.17\% improvements in F-Score on the curated blog dataset. In addition to presenting the solution to the practically useful challenge of identifying medical sentiment in a clinical context, we also significantly improve upon the highly successful classification techniques (CNN, SVM). We further evaluate our proposed approach on \textit{``2014 ShARe/CLEF task-2a (attribute normalization)''} in order to show the generic nature of our algorithm. On this task, our system achieves significantly high precision (P), recall (R), and accuracy (A) for both \textit{`Severity Class'} (P: 0.9828, R: 0.9832, A: 0.9832) and \textit{`Course Class'} (P: 0.9833, R: 0.9837, A: 0.9837) attributes. 

\begin{table*}[]
\centering
\begin{tabular}{|l|l|}
\hline
\multicolumn{1}{|c|}{\textbf{Medical Blog}} & \multicolumn{1}{c|}{\textbf{Label}} \\ \hline
\hline
\textit{\begin{tabular}[c]{@{}l@{}}``Hi \textbf{been on Sertaline} now for abut 4 weeks. My mood has\\ \textbf{definitely improved and I am alot calmer.}"\end{tabular}} & Recovered \\ \hline
\textit{\begin{tabular}[c]{@{}l@{}}``This morning I had an \textbf{attack} of it that was \textbf{very intense}.\\ I felt an \textbf{incredible surge of unsteadiness}."\end{tabular}} & Exist \\ \hline
\textit{\begin{tabular}[c]{@{}l@{}}``Had anxiety for few months on \textbf{citalopram,}  \textbf{propanolol}.\\ Nothing seems to help been in bed for two days \textbf{can’t sleep}."\end{tabular}} & Deteriorate \\ \hline
\textit{\begin{tabular}[c]{@{}l@{}}``How do you calculate FEV1\% from a FEV1 result with age \\sex and height know."\end{tabular}} & Other \\ \hline
\end{tabular}
\caption{Exemplar description of benchmark annotation scheme}
\label{annotation}\vspace{-3mm}
\end{table*}

\section{Annotation Scheme}
In this section, we define at first the benchmark setup by studying the sentiments expressed in the medical blog posts. Based on the medical sentiment, we classify the medical blog posts into the following four categories:\\
\textbf{I. Recovered:} indicates that the user has recovered from health problem and expressing the positive medical sentiment. \\
 \textbf{II. Exist:} indicates that the user is experiencing health problem and expressing the negative medical sentiment, but has not mentioned any medication.\\ 
 \textbf{III. Deteriorate:} indicates that the user's medical condition has deteriorated over the time and he/she is expressing the negative medical sentiment towards the medication.\\
 \textbf{IV. Other:} indicates that the user is discussing general topics and not expressing any medical sentiment. There is no mention of any medical symptom or medication.\\
We provide the exemplary description of the annotation scheme in Table-\ref{annotation}.

\section{Proposed Model: Network for Identifying Medical Sentiment}\label{cnn}
\begin{figure*}[h]
\includegraphics[width=14cm,height=7cm]{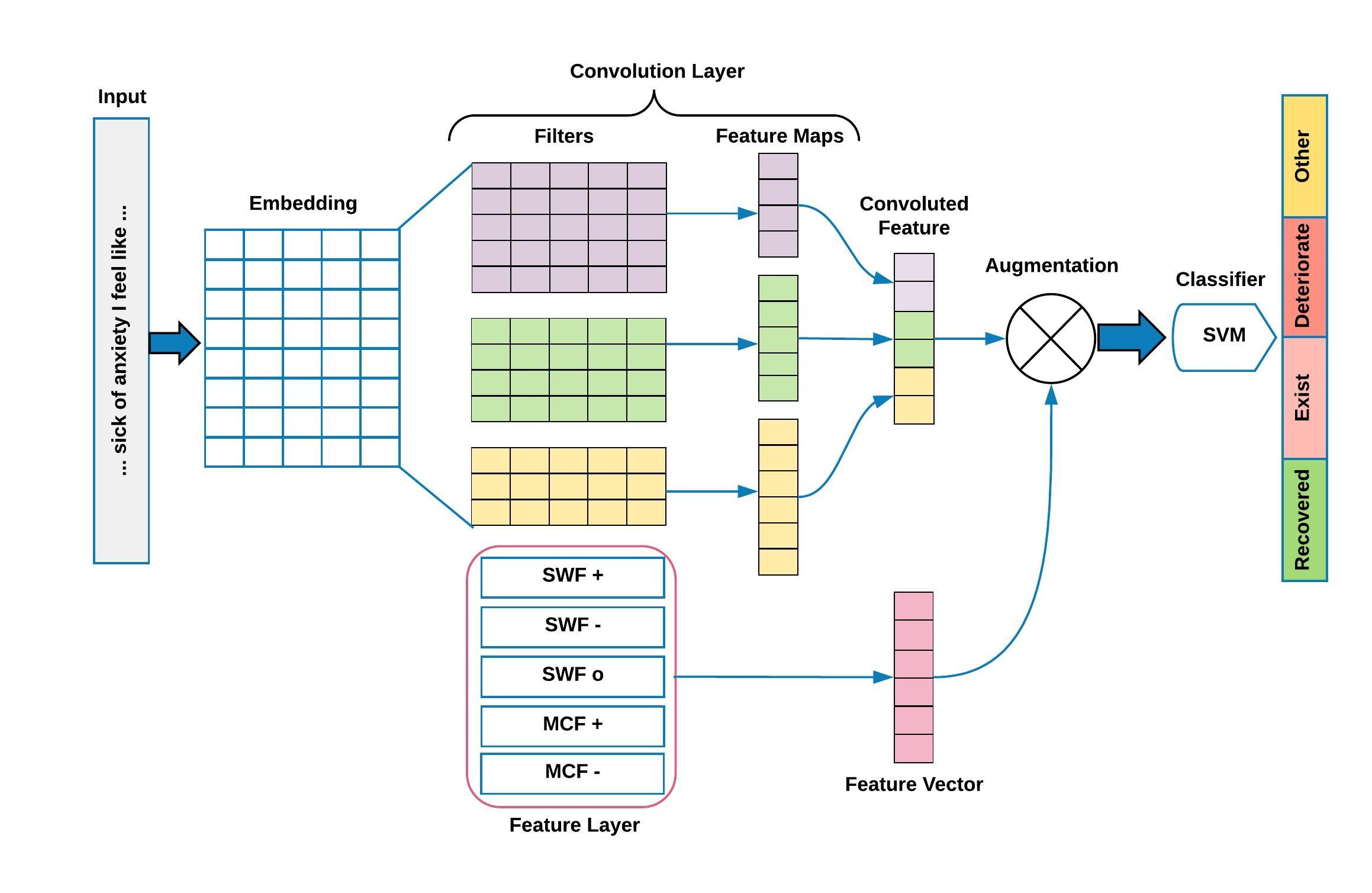}
\caption{Proposed MS-CNN architecture for predicting the m conditionmedical condition from the medical blog.}
\label{fig-model}
\end{figure*}

In this section, we propose a method based on CNN that exploits users' medical sentiments from health forums in the augmentation layer. As presented in Figure-\ref{fig-model}, the proposed model has six different components where the first four layers are similar to the conventional CNN components as proposed by \cite{kim2014convolutional}. The system takes a complete blog post as input and outputs probabilities corresponding to the four classes. We use max-pooling over the whole blog post to obtain global features through all the filters.
This pooled feature is fed into the augmentation layer instead of the fully connected neural network which concatenates pooled feature produced by the CNN with the sentiment-sensitive medical features. In the output layer, we use SVM instead of the softmax classifier to automatically classify the post into the four classes. We describe below the layers of our proposed model in details:\\
\textbf{I. Input layer:} Each blog post is provided as input to the model.\\
\textbf{II. Word embedding layer:} This layer encodes every word into a real-valued vector.
Given a blog text $M$ consisting of $n$ words ${w_1, w_2, w_3,.....w_n}$, each word $w_i$ is transformed into real-valued vector $x_i$.
Each word in $M$ is looked up in corresponding word embedding matrix $W \in R^{k \times |V|}$, where $|V|$ represents fixed-length vocabulary and $k$ is the word embedding size.
The blog-post representation matrix $x_{1:n_W}$ can be represented as:
\begin{equation}
x_{1:n_W}=x_1 \otimes x_2 \ldots \otimes x_{n_W}
\end{equation}
where $\otimes$ represents the concatenation operation.
We perform zero padding in case the number of the words in a blog text is less than $n$, to fix the length.\\
\textbf{III. Convolution layer:} Word embedding is feed as the input to the convolution layer where filter $\textbf{F}\in \mathbb{R}^{m \times k}$ is convoluted to the context window $x_{i:i+m-1}$ of $h$ words for each blog-post as follows.
\begin{equation}
c_i=f(\textbf{F}.x_{i:i+m-1} + b)
\end{equation}
where $f$ is a non-linear function\footnote{In our experiments we have used the rectified linear unit as a non linear function.} and $b$ is a bias term. The feature map $f$ is generated by applying a given filter \textbf{F} to every potential window of the words in the blog-post.
\begin{dmath}
f=[g(\textbf{F} \cdot x_{1:1+h-1} + b), g(\textbf{F} \cdot x_{2:1+h-1} + b) \ldots g( \textbf{F} \cdot x_{n-h+1:n} + b)\\
= [ f_{1}, f_{2}, f_{3} ..... f_{n-h+1}]
\end{dmath}
\vspace{-0.1em}
In order to increase the coverage of n-gram model, multiple filters with different window sizes can be applied.\\
\textbf{IV. Pooling layer:} The function of the pooling layer is to gradually minimize the spatial size of the representation by identifying the most abstracted feature generated by the convolution layer.
It involves non-linear downsampling to extract the most relevant set of the features.
In our work, we apply max-pooling operation over feature map and set the maximum value as a feature for this particular filter.
The max-pooling operation is performed over feature map as follows:
\vspace{-1em}
\begin{equation}
\hat{d}= max(f_{1}, f_{2}, f_{3} ..... f_{n-h+1})
\end{equation}
\textbf{V. Feature layer:} In this layer, we generate the sentiment features which are specific to the medical context, for each blog post as described below:\\
\indent \textbf{(1). \textit{Sentiment word feature (SWF):}} Sentiment clues words provide important features in deciding the emotions of the users. Besides, we can observe that inclusion of negation to the sentiment word can change the polarity. For example, there is positive emotion in ``I'm \textbf{stable}" but after including negation like ``I'm \textbf{not stable}", the emotion polarity changes. Briefly, there are two types of sentiment events by which we can capture the sentiments of users: occurrences of sentiment words (SW), occurrences of sentiment words with negation (NSW).
This feature calculates the positive ($SW_+$), negative ($SW_-$) and objective ($SW_O$) score for each word by capturing the sentiment event \cite{dang2009machine}. Publicly available SentiWordNet (SWN)\footnote{http://sentiwordnet.isti.cnr.it/} is used to calculate the score for each word as follows:
\begin{equation}
\begin{aligned}
SW_+=&(tf(SW)*f_{+}(SW)+tf(NSW)*f_{-}(SW))*idf(SW)
\end{aligned}
\end{equation}
\vspace{-1em}
\begin{equation}
\begin{aligned}
SW_-=&(tf(SW)*f_{-}(SW)+tf(NSW)*f_{+}(SW))*idf(SW)
\end{aligned}
\end{equation}
\vspace{-1em}
\begin{equation}
\begin{aligned}
SW_O=tfidf(SW)*f_{O}(SW)\\
\end{aligned}
\end{equation}
Here, \textit{tf} \& \textit{idf} represent the term and inverse document frequencies, respectively. $f_{+}(SW)$, $f_{-}(SW)$ and $f_{O}(SW)$ are positive, negative and objective scores, respectively obtained from the SentiWordNet. The sentiment word feature of a blog having $n$ words obtained by the following way: \\$SWF_+= \frac{\sum_{SW \in n} SW_+}{n}$ , $SWF_-= \frac{\sum_{SW \in n} SW_+}{n}$ and  $SWF_O= \frac{\sum_{SW \in n} SW_O}{n}$.\\

\indent 
\textbf{(2). \textit{Medical sentiment context feature (MCF):}} 
After analyzing the data we observe that $92$\% of the posts depict sentiments in the context of a certain stative verb such as `\textit{feel}', `\textit{suffer}', `\textit{experience}'. We design this contextual feature by considering a context window of [-$5$,$5$] words and selecting the most effective stative verb. Thereafter, negative and positive densities of a post are calculated by the frequency of the clue word to the number of words in context (i.e. $10$ in this case). For example, if a post contains more than one instance of `\textit{feel}' term, we calculate the score individually and consider the maximum one. If the word `\textit{feel}' appears at the $i^{th}$ position in a blog then the score will be calculated as follows:
\begin{equation}
\footnotesize
Score(+)=\sum_{m=i-k}^{m=i-1}w_m \times f_{+}(SW_m) + \sum_{n=i+1}^{n=i+k}w_n \times f_{+}(SW_n)
\end{equation}
\vspace{-1em}
\begin{equation}
\footnotesize
Score(-)=\sum_{m=i-k}^{m=i-1}w_m \times f_{-}(SW_m) + \sum_{n=i+1}^{n=i+k}w_n \times f_{-}(SW_n)
\end{equation}
where, $k$ is context window size, weight $w_m=m+k-i+1$ and $w_n=k-n+i+1$.
The aggregate score $MCF_+$ and $MCF_-$ of a blog post is calculated as follows:
\begin{equation}
\begin{aligned}
MCF_+=max(Score_{t=0}(+), Score_{t=1}(+), 
\ldots , Score_{t=T}(+)) 
\end{aligned}
\end{equation}
\begin{equation}
\begin{aligned}
MCF_-=max(Score_{t=0}(-), Score_{t=1}(-), 
\ldots , Score_{t=T}(-)) 
\end{aligned}
\end{equation}
where, T is the number of sentiment words in the post.
At the end the feature obtained from the pooling layer and these sentiment specific features form a augmented feature $\hat{A}$.
\begin{equation}
\begin{aligned}
\hat{A}= & \hat{d}_{i=0}  \otimes \hat{d}_{i=1} \otimes \ldots \hat{d}_{i=P} \otimes SWF_{+} \otimes SWF_{-} \otimes  SW_{O}  \otimes MCF_{+}  \otimes  MCF_{-}
\end{aligned}
\end{equation}
Where $P$ is the length of feature representation obtained from pooling layer, $\otimes$ is the concatenation operator.\\
\textbf{VI. Output layer:} The blog-level feature vector is finally passed to an SVM to perform classification. In this layer, we explore SVM instead of the traditional softmax classifier to predict the label `$y$' from a discrete set of class `$Y$=\{\textit{Recovered, Exists, Deteriorate}, and \textit{Other}\}' for the corresponding blog post `$M$'. For the multi-class classification problem, the classifiers like softmax regression (logistic regression) and SVM often provide comparable results. However, when the data is not linearly separable, SVM with non-linear kernel outperforms the Logistic Regression \cite{pochet2006support}. Furthermore, LR is prone to the problem of over-fitting as it focuses on maximizing the likelihood, while SVM can generate linear hyperplanes by mapping the data into high-dimensional spaces. This is the underlying motivation behind replacing the LR with an SVM in the final layer.




\section{Datasets, Experimental Results, and Analysis}
In this section, we present the dataset that we create and report the experimental results along with proper analysis. 

\subsection{Datasets and Experimental Setup}
We design a web crawler to collect posts of different users from the health-related forums.
We generate a corpus of $4,057$ blog posts collected from health forums \footnote{https://patient.info/, https://www.dailystrength.org/}, open social media platforms which provide evidence-based information on a variety of medical and health topics. In our study, we focused on four popular groups, namely \textit{Depression, Allergy, Asthma, and Anxiety}. Since, our main aim is to classify each post on the basis of content, we cleaned the dataset by omitting the user-names and its corresponding hyperlinks. Unlike dataset utilized in \cite{yadav2018medical,yadav2018multi}, in this dataset we have considered `Others' class because our main aim is to prioritize the individual’s posts that requires immediate attention by understanding their sentiments.

A team of three expert annotators (lexicographer with knowledge of basic medical concept) independently annotated the user posts with four classes. We use Cohen's kappa approach \cite{cohen1960coefficient} to measure the inter-annotator agreement. We observe high agreement ratio of $0.82$ for exact matching of the class w.r.t each blog post. 
The dataset statistics are shown in Table-\ref{stat}.
We perform 5-fold cross validation on the labeled dataset to learn MS-CNN model as discussed in Section-\ref{cnn}.\\ 
\begin{table}[]
\centering
\begin{tabular}{|c|c|c|c|c|}
\hline
\textbf{Dataset Statistic} & \textbf{Recovered} & \textbf{Exist} & \textbf{Deteriorate} & \textbf{Other}\\ \hline
\textit{Number of blog posts} & 136 & 1711 & 1778 & 432 \\ \hline
{\begin{tabular}[c]{@{}c@{}}\textit{Average number of sentences per post}\end{tabular}} & 5 & 5 & 7 & 3 \\ \hline
{\begin{tabular}[c]{@{}c@{}}\textit{Average number of words per post}\end{tabular}} & 152 & 137 & 186 & 19 \\ \hline
\end{tabular}
\caption{Medical blog dataset statistics}
\label{stat}
\end{table}
\textbf{CLEF eHealth Dataset:} To evaluate the effectiveness our system, we also train and test on the ``ShARe/CLEF eHealth 2014 Task 2" \cite{mowery2014task} dataset which is sampled from MIMIC-II (Multiparameter Intelligent Monitoring in Intensive Care), a database containing clinical reports of Intensive Care Unit (ICU) patients. The dataset contains $431$ clinical reports including $136$ \textit{Discharge Summary}, $54$ \textit{Radiology Report}, $54$ \textit{ECHO Report}, and $54$ \textit{ECG Report} for training, and $133$ \textit{Discharge Summary} for the test. \\
We replace each disorder mention with the keyword `DISORDER' to make our system insensitive to certain mentions and enable to focus on the context of a disorder.
Table-6 reports the corpus statistics for the attributes \textit{`Severity Class'}, and \textit{`Course Class'} and their corresponding classes.
\begin{table}[!h]
\centering
\begin{adjustbox}{width=1\textwidth}
\small
\begin{tabular}{|*{12}{c|}}  
\hline
\multirow{2}{*}{\backslashbox{\textbf{Dataset}}{\textbf{Class}}} & \multicolumn{4}{|c}{\textbf{Severity Class (\#sentences)}} & \multicolumn{7}{|c|}{\textbf{Course Class (\#sentences)}} \\\cline{2-12}
& slight & moderate & severe & unmarked & changed & increased & decreased & improved & resolved & worsened & unmarked \\ \hline
Training & 910 & 339 & 135 & 9972 & 8 & 114 & 101 & 161 & 60 & 10737 & 54 \\ \hline
Test & 234 & 189 & 76 & 7448 & 4 & 23 & 83 & 98 & 53 & 7660 & 42 \\ \hline
\end{tabular}
\end{adjustbox}
\caption{CLEF eHealth 2014 dataset statistics}
\end{table} \\
\textbf{Hyperparameter settings in SVM:} In the SVM the $C$ value \& kernel were obtained by optimizing on the development set. The value of $C$ was set to 0.01 and a Gaussian kernel was used in our experiment. A grid search was performed to deduce to an optimal value of $C$ in the range of $10^{-4}$ to $10^{2}$. We used a LibSVM implementation \footnote{https://www.csie.ntu.edu.tw/~cjlin/libsvm/} of SVM.\\
\textbf{Hyperparameter settings in CNN:}
The hyper-parameters values were determined from preliminary experiments by evaluating the model's performance on the 5-fold cross validation by varying the convolution feature sizes ($100$, $200$, \& $300$). Specifically, most of the deep learning models use the 300-dimension word embedding, the feature maps size of $300$ on the multiple filter window of size $3$, $4$, and $5$.  
We used Adam \cite{kingma2014adam} as our optimization method with a learning rate of $0.001$. 
Training was performed using stochastic gradient descent over mini-batches considering the Adadelta \cite{zeiler2012adadelta} update rule. As a regularizer, we used dropout \cite{hinton2012improving} with a probability of $0.5$. After training, we chose the best performing model to be evaluated on the test set. The MS-CNN model introduced in this paper is implemented in Theano \footnote{http://deeplearning.net/software/theano/}.
\vspace{-2mm}
\subsection{Result and Analysis}
\vspace{-2mm}
In order to make an effective comparison of our proposed approach, we design two strong baselines as following:\\
\textbf{(1) Baseline 1:} The first baseline model is constructed by training an SVM with Bag-of-Words (BoW) and the sentiment-sensitive features presented in the feature layer of MS-CNN in Figure 1.\\
\textbf{(2) Baseline 2:} In this model, we use the standard CNN model learned using only word embedding features.\\
\begin{table}[h]
\centering
\begin{adjustbox}{width=1\textwidth}
\small
\begin{tabular}{llll}
\hline
\textbf{System} & \textbf{Precision} & \textbf{Recall} & \textbf{F-Score} \\ \hline
\textbf{Baseline-1 \textit{(BoW+SVM)}} & 0.7218 & 0.7347 & 0.7281 \\ \hline
\textbf{Baseline-2 \textit{(CNN)}} & 0.8204 & 0.7917& 0.8057 \\ \hline
\textbf{\textit{CNN (Pooling)+SVM}} & 0.8254 &0.8119 & 0.8185 \\ \hline
\textbf{\textit{MS-CNN (Pooling+SWF+MCF+SVM)}} & 0.8305  & 0.8499 & 0.8393 \\ \hline
\end{tabular}
\end{adjustbox}
\caption {Performance comparison of our proposed approach (MS-CNN) with baselines on the medical blog dataset.}
\label{RESULT-CNN} 
\end{table}
Table-\ref{RESULT-CNN} reports the performance of our proposed approach (MS-CNN) with other baselines where we observe improvements of $15.27\%$ and $4.17\%$ over Baseline $1$ and $2$, respectively. Statistical significance test shows that improvement over both the baselines are significant (\textit{p-value} $<$ $0.05$).\\
We perform experiments to select the best filter length using all the features. We perform 5-fold cross validation by varying the convolution feature sizes ($100$, $200$ \& $300$) as reported in Table-\ref{filtewidth}. It is observed that increase in the feature size generally tends to enhance the performance. We obtain the best performance of $83.93$\% F-Score using the feature size of $300$.
Providing small window size (filter length) often fails to capture sufficient context, whereas too long tends to include irrelevant contextual information. Based on these observations, we set the filter lengths to a minimum $3$ and maximum of $6$. Further, we analyze the influence of multiple filters on our evaluation. Results indicate that combination of three filters is the better choice to improve the performance. For example, the combination of filter lengths $3$, $4$ and $5$ provide the highest recall value of $84.99$ \% on $300$ feature size. We also observe the similar behavior on $100$ and $200$ feature sizes. Any further increase in filter length reduces the performance. For example, the combination of filter lengths $3$, $4$, $5$ and $6$ does not lead to performance improvement.
\begin{table*}[h]
\centering
\resizebox{0.7\textwidth}{!}{
\begin{tabular}{c|c|c|c|c}
\hline
\textbf{Convolution Feature Size} & \textbf{Window Size} & \textbf{Precision} & \textbf{Recall} & \textbf{F-Score}   \\ \hline
$100$ & 3,4 & $0.7626$  & $0.7860$ & $0.7712$ \\ \hline
$100$ & 4,5 & $0.7832$ &  $0.7921$ & $0.7784$ \\ \hline
$100$ & 3,4,5 & $0.8092$ & $0.7983$ & $0.7996$ \\ \hline
$100$ &3,4,5,6 & $0.7848$ & $0.7934$ & $0.7797$  \\ \hline
$200$ & 3,4 & $0.7968$  & $0.8130$  & $0.8043$  \\ \hline
$200$ & 4,5 & $0.7984$  & $0.8093$  & $0.8024$  \\ \hline
$200$ & 3,4,5 & $0.8029$   &  $0.8229$  & $0.8099$  \\ \hline
$200$ & 3,4,5,6 & $0.8109$ & $0.8303$ &  $0.8187$  \\ \hline
$300$ & 3,4 &  $0.8252$ &  $0.8450$ & $0.8341$  \\ \hline
$300$ & 4,5 &$0.8265 $  & $0.8462$  & $0.8353$ \\ \hline
$300$ & 3,4,5 & $0.8305$ & $0.8499$ & $0.8393$ \\ \hline
$300$ & 3,4,5,6 & $0.8214$ & $0.8401$ & $0.8297$ \\ \hline
\end{tabular}
}
\caption{Performance of MS-CNN (Pooling+SWF+MCF+SVM) with different convolution feature and window sizes on the medical blog dataset}
\label{filtewidth}
\end{table*}
In order to investigate the contribution of the features i.e. Sentiment word feature (SWF) \& Medical sentiment context feature (MCF), we incorporate one feature at a time in our proposed model sequentially and observe its effect. Table-\ref{impact} provides the details of incorporating features into CNN model. We observe that incorporation of MCF feature in CNN-SVM architecture improves the performance by $1.06$\% F-score. Addition of SWF feature leads to $1.75$\% increase in F-score. Addition of both MCF \& SWF further improve recall, precision and F-score values by $4.68\%$, $0.61\%$ and $2.54$\% respectively.
In the second phase of our experiment, we include the feature to the basic CNN model learned on logistic regression classifier. 
The evaluation shows that 
SWF feature has more influence in improving the overall performance. \\
We further perform experiments with different classifiers like Ridge (RC) \cite{Hoerl:2000:RRB:338441.338461}, Logistic Regression (LR), Nearest Neighbors (NN) \cite{cover1967nearest}, Multi-layer Perceptron (MLP) \cite{gardner1998artificial} at the output layer of CNN architecture. We observe that SVM performs superior compared to the other classifiers as presented in Table-\ref{RESULT-other-classifier}. An overall increment of 2.9\% F-score is observed with the use of SVM replacing softmax in the output layer of CNN.\\ 
In the CLEF dataset, the attribute modifiers are explicit and appear mostly within a context window of $5$. For example, in the sentences ``The patient was also noted to have a \textit{significant} \underline{perineal injury}", and
``The patient's \underline{diarrhea} \textit{continued}", \textit{significant} is a severity modifier (severe), and \textit{continued} is a course class modifier (increased) for the disorder mentions \underline{perineal injury}, and \underline{diarrhea} respectively.\\
As the dataset is fairly clean, structured, and the sentences are short in length compared to the blog posts, we could achieve a significantly high accuracy of 0.9832 and 0.9837 for identifying the \textit{Severity Class}, and \textit{Course Class} of a disorder mention which outperforms the state-of-the-art system as presented in Table \ref{CLEF-comparison}. 

\begin{table}[h]
\begin{minipage}{0.45\textwidth}
\centering
\begin{adjustbox}{width=1\textwidth}
\small
\begin{tabular}{llll}
\hline
\textbf{Systems} & \textbf{Precision} & \textbf{Recall} & \textbf{F-Score} \\ \hline
\textit{Pooled Feature+SWF+LR} & 0.8219 & 0.7984 & 0.8099 \\ \hline
\textit{Pooled Feature+MCF+LR} & 0.8192 &0.7981 & 0.8085 \\ \hline
\textit{Pooled Feature+SWF+MCF+LR} &0.8074  &  0.8290 & 0.8153 \\ \hline
\textit{Pooled Feature+SWF+SVM} & 0.8371 &0.8289 &0.8329 \\ \hline
\textit{Pooled Feature+MCF+SVM} & 0.8311&0.8235& 0.8272 \\ \hline
\textit{Pooled Feature+SWF+MCF+SVM} & 0.8305  & 0.8499 & 0.8393 \\ \hline
\end{tabular}
\end{adjustbox}
\caption{Impact of each feature on performance with different classifiers on the medical blog dataset.}
\label{impact}
\end{minipage}
~\hfill~
\begin{minipage}{0.55\textwidth}
\centering
\begin{adjustbox}{width=1\textwidth}
\small
\begin{tabular}{llll}
\hline
\textbf{Systems} & \textbf{Precision} & \textbf{Recall} & \textbf{F-Score} \\ \hline
\textit{MS-CNN(Pooling+SWF+MCF)+Ridge Classifier} & 0.7615  &  0.7835 & 0.7701\\ \hline
\textit{MS-CNN(Pooling+SWF+MCF)+Logistic Regression} & 0.8074  &  0.8290 & 0.8153 \\ \hline
\textit{MS-CNN(Pooling+SWF+MCF)+ Nearest Neighbors} & 0.7796 & 0.7860 & 0.7712\\ \hline
\textit{MS-CNN(Pooling+SWF+MCF)+ MLP} & 0.7901 & 0.7983 & 0.7893  \\ \hline
\end{tabular}
\end{adjustbox}
\caption{Performance comparison of our proposed approach (MS-CNN) with other classifiers on the medical blog dataset.}
\label{RESULT-other-classifier}
\end{minipage}
\begin{center}
\vspace{-6mm}
\end{center}
\end{table}

\begin{table}[!h]
\centering
\begin{adjustbox}{width=1\textwidth}
\small
\begin{tabular}{|*{9}{c|}}  
\hline
\textbf{System} & \multicolumn{4}{|c}{\textbf{Severity Class}} & \multicolumn{4}{|c|}{\textbf{Course Class}} \\\cline{2-9}
& Precision & Recall & F-score & Accuracy & Precision & Recall & F-score & Accuracy \\ \hline
\textbf{MS-CNN} & \textbf{0.9828} & \textbf{0.9832} & \textbf{0.9829} & \textbf{0.9832} & \textbf{0.9833} & \textbf{0.9837} & \textbf{0.9833} & \textbf{0.9837} \\ \hline
TeamHITACHI\cite{johri2014optimizing} & - & - & - & 0.982 & - & - & - & 0.971 \\ \hline
RelAgent\cite{ramanancocoa} & - & - & - & 0.975 & - & - & - & 0.970 \\ \hline
\end{tabular}
\end{adjustbox}
\caption{Comparison with the state-of-the-art systems on \textit{``2014 ShARe/CLEF task-2a (attribute normalization)''} for \textit{`Severity Class'}, and \textit{`Course Class'}}
\label{CLEF-comparison}\vspace{-4mm}
\end{table}
\vspace{-3mm}
\subsection{Error Analysis}
In this section, we analyze different sources of errors that lead to misclassification. 
We closely study the false positives and false negatives and categorize the errors in the following classes: \\ 
\textbf{(1) Short blog text:} We observe that our system is unable to identify the appropriate class for the short text despite of having explicit sentiment-bearing words in it. For e.g \textit{``Had bad pains since this morning more like a shooting/stabbing pain!.''}. In this text despite of having mention of the term \textit{'pain'} explicitly, the system classifies it into \textit{'Other'} class. This might be because of the zero padding performed during pre-processing of SM-CNN. One more possible reason could be because majority of instances in `Other' class have correlation with the short texts (average three sentences). We observe that approximately $26$\% of the total errors are due to the shorter blog-text.\\ 
\textbf{(2) Implicit sentiments:} Our system fails to identify the class where sentiments are often contained implicitly. In our analysis, we observe that the words used in the medical blogs to express sentiments greatly differ from other social media. In non-medical social media, polarities are manifested in corresponding words (mainly adjectives) while in medical domain, sentiments are often presented implicitly which need to be inferred, for instance, from the medical concepts used in documents. Implicit descriptions of medical conditions are for e.g \textit{severe pain, tight chest, rapid weight gain,} etc.
We identify that $61$\% errors occur due to the problems of implicit sentiments.
For e.g, in the text, \textit{``Why does anxiety feels like you have to make yourself breathe instead of letting your body breathe on its own.!!''}, it is implicit that the user is suffering from shortness of breathe.\\
\textbf{(3) Presence of abbreviated and short words:} We analyze that the system misclassifies in presence of abbreviated word forms or short words. For example, \textit{`Cit'} instead of \textit{`Citopram'}, \textit{`CBT'} instead of \textit{'Cognitive behavioral therapy'}, \textit{`ECG'} instead of \textit{`electrocardiogram'}. The system also misclassifies when words are misspelled, for example, \textit{xanac} instead of the drug \textit{Xanax}. We analyze that approximately $13$\% of texts are misclassified due to the presence of abbreviated and short words.

\section{Conclusion}
In this paper, we analyze different aspects of medical sentiments to identify the fine-grained conditions of patients from medical blog posts. Utilize highly representative medical blogs we validate our study. We create a benchmark setup by crawling relevant data, defining classification tagset, and manually annotating them. We propose a robust sentiment-sensitive deep learning model leveraging the medical sentiment features for classifying the medical blog posts. Our experiments show that augmenting sentiment features to CNN derived features is an effective way to combine the two informations severity classification performance. In future, we would like to explore the deep learning techniques to capture implicit sentiments. 
\bibliographystyle{acl}
\bibliography{coling2018}

\end{document}